\documentclass[letterpaper]{article} 
\usepackage{aaai25}  
\usepackage{times}  
\usepackage{helvet}  
\usepackage{courier}  
\usepackage[hyphens]{url}  
\usepackage{graphicx} 
\usepackage{natbib}  
\usepackage{caption} 
\usepackage{algorithm}
\usepackage{algorithmic}

\usepackage{multirow}
\usepackage{amsthm,amsmath,amssymb}

\usepackage{mathrsfs}

\usepackage{bbding}
\usepackage{subfigure}
\usepackage{makecell}
\usepackage[table]{xcolor}
\definecolor{light-gray}{gray}{0.88}

\usepackage{newfloat}
\usepackage{listings}
\DeclareCaptionStyle{ruled}{labelfont=normalfont,labelsep=colon,strut=off} 
\floatstyle{ruled}
\newfloat{listing}{tb}{lst}{}
\floatname{listing}{Listing}
\title{Relieving Universal Label Noise for Unsupervised Visible-Infrared Person Re-Identification by Inferring from Neighbors}
\author{
    Xiao Teng\textsuperscript{\rm 1}\equalcontrib,
    Long Lan\textsuperscript{\rm 1}\equalcontrib,
    Dingyao Chen\textsuperscript{\rm 1},
    Kele Xu\textsuperscript{\rm 1},
    Nan Yin\textsuperscript{\rm 2}\thanks{Nan Yin is the corresponding author.}
}
\affiliations{
    \textsuperscript{\rm 1}College of Computer, National University of Defense Technology\\
    \textsuperscript{\rm 2}Hong Kong University of Science and Technology

    \{tengxiao14,long.lan,chendingyao\}@nudt.edu.cn, \{kelele.xu,yinnan8911\}@gmail.com

}

\usepackage{bibentry}

\begin{document}

\maketitle

\begin{abstract}
Unsupervised visible-infrared person re-identification (USL-VI-ReID) is of great research and practical significance yet remains challenging due to the absence of annotations. Existing approaches aim to learn modality-invariant representations in an unsupervised setting. However, these methods often encounter label noise within and across modalities due to suboptimal clustering results and considerable modality discrepancies, which impedes effective training. To address these challenges, we propose a straightforward yet effective solution for USL-VI-ReID by mitigating universal label noise using neighbor information. Specifically, we introduce the Neighbor-guided Universal Label Calibration (N-ULC) module, which replaces explicit hard pseudo labels in both homogeneous and heterogeneous spaces with soft labels derived from neighboring samples to reduce label noise. Additionally, we present the Neighbor-guided Dynamic Weighting (N-DW) module to enhance training stability by minimizing the influence of unreliable samples. Extensive experiments on the RegDB and SYSU-MM01 datasets demonstrate that our method outperforms existing USL-VI-ReID approaches, despite its simplicity. The source code is available at: https://github.com/tengxiao14/Neighbor-guided-USL-VI-ReID.
\end{abstract}

%

\section{Introduction}
Visible-infrared person re-identification (VI-ReID) focuses on identifying the same person from the visible or infrared camera when a query image is provided from the other modality \cite{hao2021cross, ye2021channel}. It has attracted widespread attention due to its potential in night vision surveillance applications, where traditional visible cameras cannot work well. Benefiting from accurate manual annotations, recent works have achieved notable improvement on the VI-ReID task. However, the reliance on annotations within and across modalities heavily limits their applications. To address the issue, the task of unsupervised visible-infrared person re-identification (USL-VI-ReID) has emerged and garnered substantial interest \cite{yang2022augmented, yang2024shallow}. 

\begin{figure}[t] 
\centering
\includegraphics[width=0.47\textwidth,height=7.3cm]{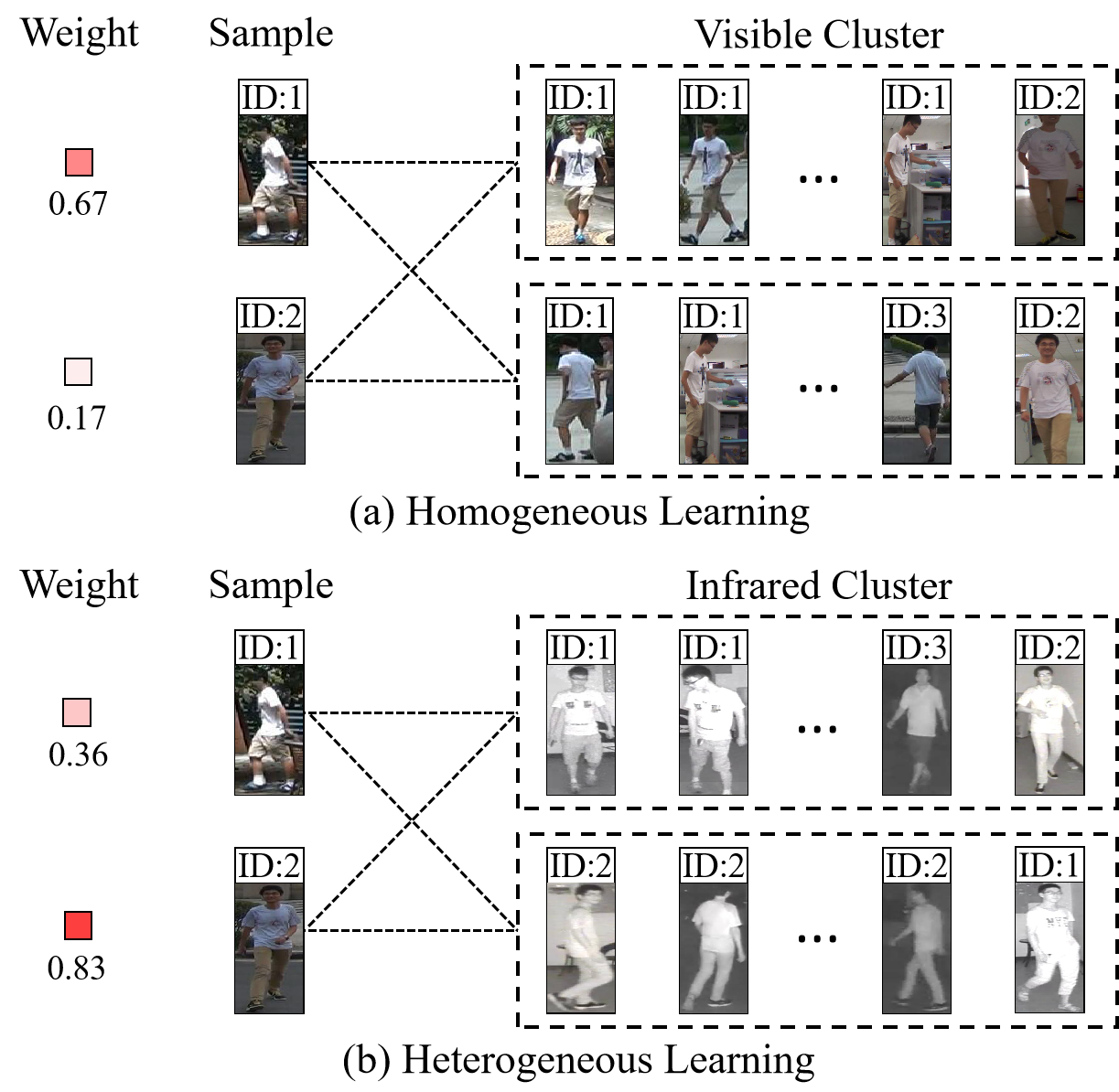}
\caption{Illustration of the motivation of our method. Due to inferior clustering results, images of the same person can be divided into different groups in each modality. As a result, one-hot pseudo labels are limited to represent their identities in both homogeneous and heterogeneous learning processes. Additionally, the reliability of labeling can vary for diverse samples during the learning process.
}\label{illustration}
\end{figure}

The key of USL-VI-ReID is to learn  modality-shareable feature representations using only an unlabeled training set \cite{liang2021homogeneous}. To achieve this, existing approaches \cite{yin2024sport, cheng2023unsupervised, cheng2023efficient} typically generate pseudo labels to optimize model parameters. Specifically, these methods first apply clustering algorithms to create pseudo labels within each modality independently. Subsequently, these pseudo labels are used to establish cross-modality correspondences, which aim to create a definitive association between visible and infrared modalities. However, as illustrated in Fig. \ref{illustration}, suboptimal clustering results can lead to images of the same person being assigned to different groups. Consequently, relying solely on one-hot pseudo labels can introduce significant label noise both within and across modalities, thus impeding the training process.  

To address the issue, we propose a simple yet effective framework for USL-VI-ReID by relieving universal label noise with neighbor information. The motivation of our method is that the real identity of a query image can be inferred from its neighbors within and across modalities. Initially, recognizing that images of the same person can be divided into groups and one-hot vectors are limited to represent their identities, we propose the neighbor-guided universal label calibration module to replace the explicit one-hot pseudo labels within and across modalities with more realistic soft pseudo labels derived from the pseudo labels of neighboring images. This approach alleviates universal label noise by providing more accurate representations of true identities. Furthermore, to enhance training stability, we introduce the neighbor-guided dynamic weighting module, which reduces the impact of unreliable samples in both homogeneous and heterogeneous learning processes by utilizing the consistency of neighbors' pseudo labels. By integrating these modules, we significantly reduce label noise at both the pseudo label and sample levels. Our contributions can be summarized as follows:
\begin{enumerate}
\item[$\bullet$] We propose a straightforward yet efficient framework for USL-VI-ReID that utilizes neighbor information to alleviate universal label noise within and across modalities at both the pseudo label and sample levels.
\item[$\bullet$] We propose the neighbor-guided label calibration module, which aims to provide more accurate identity representation by employing realistic soft labels for images within and across modalities.
\item[$\bullet$] We further introduce the neighbor-guided dynamic weighting module to diminish the influence of unreliable samples by utilizing the consistency of neighboring pseudo labels in both homogeneous and heterogeneous learning contexts.
\item[$\bullet$] Extensive experiments conducted on two public visible-infrared person ReID benchmarks demonstrate that our method can outperform existing approaches across various settings.
\end{enumerate}

\section{Related Work}
\subsection{Supervised Visible-Infrared Person ReID}
Supervised visible-infrared person ReID (SVI-ReID) aims to match the same person across different modalities. The key of SVI-ReID is to obtain discriminative feature representations by leveraging accurate manual annotations \cite{wu2017rgb, Kim_2023_CVPR}. Among them, Wu \textit{et al.} \cite{wu2017rgb} introduce the task of SVI-ReID and proposes a zero-padding approach to boost the performance of one-stream model. To regularize the model from overfitting, the PartMix augmentation technique \cite{Kim_2023_CVPR} is introduced by mixing local descriptions of images from different modalities for part-level data augmentation. To enhance the generalization of graph-based models, Li \textit{et al.} \cite{li2022counterfactual} rethink the SVI-ReID problem from the perspective of counterfactual intervention and propose a novel framework by transferring features and stressing the effect of topology structure. Regarding that convolution neural network is limited in extracting discriminative feature representations, Zhao \textit{et al.} introduce the transformer into the task of SVI-ReID and propose a novel framework by enhancing spatial and channel information. 

While these methodologies have demonstrated substantial efficacy in learning modality-shareable feature representations, they all require expensive annotations that are often challenging to acquire, thereby constraining their real-world applicability.

\subsection{Unsupervised Visible-Infrared Person ReID}
Unsupervised visible-infrared person re-identification (USL-VI-ReID) seeks to learn modality-shareable feature representations without requiring annotations \cite{yang2022augmented, wu2023unsupervised,pang2024mimr}. This task is inherently more challenging than unsupervised single-modality person ReID, as the discrepancies between modalities often surpass the intra-class variance within a single modality \cite{teng2023highly, lan2023learning, zhao2022heterogeneous}. To effectively utilize unlabeled training data from diverse modalities, current approaches typically establish cross-modality correspondences through clustering within each modality \cite{ju2024survey, wu2023unsupervised, pang2023cross}. For instance, Yang \textit{et al.} \cite{yang2022augmented} introduce a novel contrastive learning framework that aggregates cross-modality memories based on priority counts. Wu \textit{et al.} \cite{wu2023unsupervised} frame cross-modality correspondence mining as a graph matching problem and propose a method featuring progressive graph matching and alternative cross-modality learning modules. 

Despite these advancements, the methods still face challenges due to suboptimal clustering results and significant modality discrepancies, which can introduce label noise both within and across modalities, thereby limiting model performance.

\subsection{Neighbor-Related Person ReID}
Since a single image contains limited information, recent studies have sought to extract more valuable information from neighboring images in both supervised and unsupervised settings \cite{wang2022nformer, zhong2017re, yu2021neighbourhood}. For instance, Wang \textit{et al.} \cite{wang2022nformer} introduce explicit relationships between input images to mitigate outlier effects and enhance the robustness of learned feature representations. Zhong \textit{et al.} \cite{zhong2017re} utilize neighbor information to refine retrieval results derived from direct similarity calculations of feature vectors. To address occlusion issues in person ReID, Yu \textit{et al.} \cite{yu2021neighbourhood} propose a neighbor-guided feature reconstruction method to recover missing information using reliable neighbor data

Similar to our approach, several studies also incorporate neighbor information in USL-VI-ReID tasks as a sub-module to refine cross-modality correspondences or to extract relationships among samples \cite{cheng2023unsupervised, pang2023cross, he2024exploring, yin2024robust, yang2023dual}. However, our method is distinct in that it is exclusively based on neighbor information and employs it in a more comprehensive and universal manner. This approach addresses pseudo label noise in both homogeneous and heterogeneous spaces through pseudo label refinement and sample weighting. As a result, our method demonstrates superior efficiency compared to existing methods, despite its inherent simplicity. 

\begin{figure*}[t] 
\centering
\includegraphics[width=1.0\textwidth,height=9.6cm]{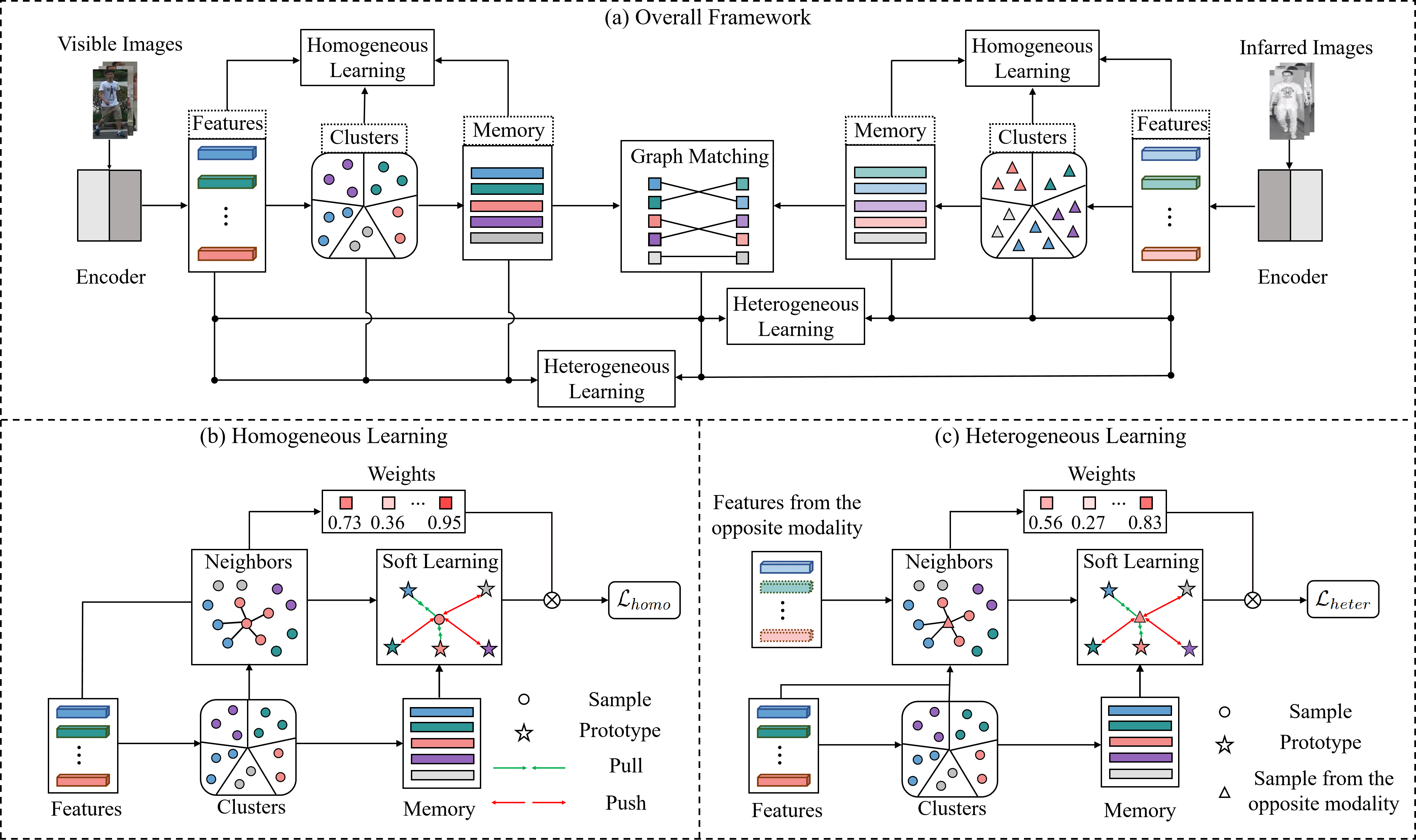}
\caption{\textbf{Framework of the proposed method.} Based on the Progressive Graph Matching (PGM) framework, we propose the neighbor-guided universal label calibration module (Sec.~\ref{calibration}) and neighbor-guided dynamic weighting module (Sec.~\ref{weighting}), these module are applied on both the (b) homogeneous learning and (c) heterogeneous learning processes.}\label{fig2}
\end{figure*}

\section{Method}
The framework of our proposed method is depicted in Fig. \ref{fig2}. To provide a comprehensive understanding, we first revisit the Progressive Graph Matching (PGM) framework \cite{wu2023unsupervised}, which serves as the baseline for our approach. Building upon PGM, we will then introduce two key components of our method: the Neighbor-guided Universal Label Calibration (N-ULC) module and the Neighbor-guided Dynamic Weighting (N-DW) module, which will be detailed in the subsequent sections.

\subsection{Progressive Graph Matching}
\label{baseline}
Given the unlabeled training sets for visible and infrared modalities, denoted as $X=\left\{X_{v}, X_{i} \right\}$, where $X_{v}=\left\{x_{1}^{v}, x_{2}^{v}, \ldots, x_{N}^{v}\right\}$ represents the visible training set with $N$ images, while $X_{i}=\left\{x_{1}^{i}, x_{2}^{i}, \ldots, x_{M}^{i}\right\}$ denotes the infrared training set containing $M$ images. To extract features from the unlabeled training sets, the two-stream encoders $f_{\theta}^{v}$ and $f_{\theta}^{i}$ are utilized, which share the same convolution backbone but with modality-specific classifiers. This process yields modality-specific feature sets $U_{v}=\left\{u_{1}^{v}, u_{2}^{v}, \ldots, u_{N}^{v}\right\}$ and $U_{i}=\left\{u_{1}^{i}, u_{2}^{i}, \ldots, u_{M}^{i}\right\}$ for visible and infrared modalities, respectively. To generate pseudo labels for these feature sets, we apply the DBSCAN clustering algorithm \cite{ester1996density} independently on $U_{v}$ and $U_{i}$. This results in cluster sets $\mathcal{H}_{v}=\left\{C_{1}^{v}, C_{2}^{v}, \ldots, C_{K}^{v}\right\}$ and $\mathcal{H}_{i}=\left\{C_{1}^{i}, C_{2}^{i}, \ldots, C_{L}^{i}\right\}$, where $K$ and $L$ denote the number of clusters in the visible and infrared modalities, respectively. Then the modalities-specific prototypes $\Phi_{v}\in\mathbb{R}^{K\times d}$ and $\Phi_{i}\in\mathbb{R}^{L\times d}$ can be obtained by applying averaging operation on feature vectors within each cluster, where d denotes the dimension of the feature vector.

To establish associations between different modalities, the graph matching algorithm can be formulated as follows: 
\begin{equation}\label{PGM}
\begin{aligned}
G(M) = \mathop{\arg\min}\limits_{M} D^{T}M \\
s.t. \forall i\in [K], j\in [L]: M_{ij}\in\{0,1\},\\
\forall i\in [K]: \sum\limits_{j\in [L]}{M_{ij}\leq 1},\\
\forall j\in [L]: \sum\limits_{i\in [K]}{M_{ij}\leq 1},
\end{aligned}
\end{equation}
where $D\in\mathbb{R}^{K\times L}$ denotes the cost matrix representing the dissimilarity measure between the visible prototype $\Phi_{v}\in\mathbb{R}^{K\times d}$ and infrared prototype $\Phi_{i}\in\mathbb{R}^{L\times d}$. Using the progressive matching strategy detailed in \cite{wu2023unsupervised}, it is possible to associate all clusters between the visible and infrared modalities. Due to space constraints, additional details on the matching process are omitted here. Consequently, the cross-modality label transformers $T^{V\rightarrow I}$ and $T^{I\rightarrow V}$ are derived, facilitating the transformation of pseudo labels between modalities. During training, for a sample $q_v$ from visible modality, assumed to belong to the $k'$-th cluster, homogeneous learning can be expressed as follows:
\begin{equation}\label{PGM-intra}
L^{intra}({q}_{v})=-\log \frac{\exp \left({q}_{v} \cdot \phi_{k'}^{v} / \tau\right)}{\sum_{k=1}^{K} \exp \left({q}_{v} \cdot \phi_{k}^{v} / \tau\right)},
\end{equation}
where $\phi_{k}^{i}$ denotes the prototype of the $k$-th cluster from visible modality, while $\phi_{+}^{v}$ denotes the prototype of the cluster to which $q_v$ belongs. $\tau$ is the temperature hyper-parameter. Prototypes are updated using a momentum scheme based on their corresponding query features \cite{wu2023unsupervised}. Similarly, heterogeneous learning can be formulated as follows:
\begin{equation}\label{PGM-inter}
L^{inter}({q}_{v})=-\log \frac{\exp \left({q}_{v} \cdot \phi_{T^{V\rightarrow I}[k']}^{i} / \tau\right)}{\sum_{l=1}^{L} \exp \left({q}_{v} \cdot \phi_{l}^{i} / \tau\right)},
\end{equation}
where $\phi_{l}^{i}$ denotes the $l$-th cluster in the infrared modality. To mitigate the modality gap, Random Channel Augmentation (CA) \cite{ye2021channel} is employed during training. For samples from the infrared modality, comparisons are made with both infrared and visible prototypes in a similar manner. The objective function is then formulated as follows:
\begin{equation}\label{PGM-learning}
\begin{aligned}
L_{overall} =  \frac{1}{|B_V|} \sum_{q_v \in B_V}(L^{intra}({q}_{v}) + \lambda L^{inter}({q}_{v})) \\
+ \frac{1}{|B_I|} \sum_{q_i \in B_I}(L^{intra}({q}_{i}) + \lambda L^{inter}({q}_{i})),
\end{aligned}
\end{equation}
where $\lambda$ is the hyper-parameter that balances homogeneous learning and heterogeneous learning.

\subsection{Neighbor-Guided Universal Label Calibration}
\label{calibration}
Due to suboptimal clustering results and significant modality discrepancies, learning with a progressive graph matching framework can be adversely affected by substantial label noise both within and across modalities. Since images of the same individual may be assigned to different clusters during the clustering process, relying solely on hard pseudo labels to establish connections between images and a single cluster, either within or across modalities, proves inadequate. To address this challenge, we propose the N-ULC module. This module aims to enhance the accuracy of sample identity representation by deriving soft labels from information provided by neighboring samples.

Specifically, given the training sets from different modalities, we obtain feature sets $\{U_{v}, U_{i}\}$ and cluster sets $\{\mathcal{H}_{v}, \mathcal{H}_{i}\}$ as described in the previous section. For each sample $q_v$ from the visible training set, which is assumed to belong to the $l'$-th cluster, we identify its $k$-nearest neighbors from the visible training set $U_v$. The resulting list of neighbors is denoted as $N(q_v, U_{v}, k)$. Intuitively, the identity statistics of this ranking list can provide insights into the true identity of $q_v$. Consequently, the correlation between $q_v$ and the visible clusters can be expressed as: 
\begin{equation}\label{correlation}
[\tilde{P}_{q_v}^{intra}]_l =  \frac{|N(q_v, U_{v}, k)\cap C^v_l|}{|N(q_v, U_{v}, k)\cup C^v_l|},
\end{equation}
where $\tilde{P}_{q_v}^{intra}$ represents the correlation between $q_v$ and the visible clusters, and $[\tilde{P}_{q_v}^{intra}]_l$ denotes the $l$-th entry of $\tilde{P}_{q_v}^{intra}$. Since the query sample $q_v$ belongs to the $l'$-th cluster in the visible modality, its original one-hot pseudo label is denoted as $\mathcal{I}_{q_v}^{intra}$, where the $l'$-th entry is set to 1 and all other entries are set to 0. The calibrated soft label for $q_v$ can then be expressed as:
\begin{equation}\label{calibrate_intra}
\tilde{\mathcal{I}}_{q_v}^{intra} =  \mu \mathcal{I}_{q_v}^{intra} + (1-\mu){P}_{q_v}^{intra},
\end{equation}
where ${P}_{q_v}^{intra}$ denotes the $\ell_1$-norm of $\tilde{P}_{q_v}^{intra}$ and $\mu$ is the hyper-parameter. Drawing inspiration from the label smoothing technique, the calibrated homogeneous learning process for the query sample $q_v$ can be expressed as follows:
\begin{equation}\label{soft-intra}
L^{intra}_{soft}({q}_{v})=-\sum\limits_{k=1}\limits^{K}[\tilde{\mathcal{I}}_{q_v}^{intra}]_k \log \frac{\exp \left({q}_{v} \cdot \phi_{k}^{v} / \tau\right)}{\sum_{k'=1}^{K} \exp \left({q}_{v} \cdot \phi_{k'}^{v} / \tau\right)}.
\end{equation}

Similarly, label noise across modalities can also be mitigated using this approach. For a given query sample $q_v$, its cross-modality $k$-nearest neighbors can be identified as $N(q_v, U_i, k)$. Subsequently, the correlation between $q_v$ and clusters from the infrared modality can be represented as:
\begin{equation}\label{correlation2}
[\tilde{P}_{q_v}^{inter}]_l =  \frac{|N(q_v, U_{i}, k)\cap C^i_l|}{|N(q_v, U_{i}, k)\cup C^i_l|},
\end{equation}
where $U_i$ denotes the feature set extracted from the infrared training data. $C^i_l$ represents the cluster set containing samples from the $l$-th cluster in the infrared modality. Since $q_v$ belongs to the $l'$-th cluster in the visible modality, its corresponding cluster in the infrared modality can be identified using the cross-modality label transformer $T^{V\rightarrow I}$, denoted as $T^{V\rightarrow I}[l']$. Consequently, the cross-modality one-hot pseudo label for $q_v$ is represented as $\mathcal{I}_{q_v}^{inter}$, where the $T^{V\rightarrow I}[l']$-th entry is set to 1 and all other entries are set to 0. Thus, the calibrated cross-modality soft label for $q_v$ is expressed as:
\begin{equation}\label{calibrate_inter}
\tilde{\mathcal{I}}_{q_v}^{inter} =  \mu \mathcal{I}_{q_v}^{inter} + (1-\mu){P}_{q_v}^{inter},
\end{equation}
where ${P}_{q_v}^{inter}$ is the $\ell_1$-norm of $\tilde{P}_{q_v}^{inter}$. The calibrated heterogeneous learning process for $q_v$ can then be described as:
\begin{equation}\label{soft-inter}
L^{inter}_{soft}({q}_{v})=-\sum\limits_{k=1}\limits^{L}[\tilde{\mathcal{I}}_{q_v}^{inter}]_k \log \frac{\exp \left({q}_{v} \cdot \phi_{k}^{i} / \tau\right)}{\sum_{k'=1}^{L} \exp \left({q}_{v} \cdot \phi_{k'}^{i} / \tau\right)}.
\end{equation}

The same procedure can be applied to samples from the infrared modality for label calibration both within and across modalities. For simplicity, the details of this process are omitted.

\subsection{Neighbor-Guided Dynamic Weighting}
\label{weighting}
Our proposed neighbor-guided universal label calibration alleviates label noise by more accurately reflecting the true identity of images both within and across modalities~\cite{yin2023omg}. To enhance the stability of the learning process, we introduce the N-DW module, which reduces the impact of unreliable samples during training. Specifically, for a query sample $q_v$ from the $l'$-th cluster in visible modality, its normalized correlations with clusters within and across modalities can be represented as ${P}_{q_v}^{intra}$ and ${P}_{q_v}^{inter}$, respectively. Intuitively, samples that are more consistent with their neighbors are considered to have more reliable pseudo labels. Consequently, neighbor information is utilized to generate weights for samples, thereby minimizing the influence of unreliable samples. For homogeneous learning, the weight of $q_v$ is computed as follows:
\begin{equation}\label{weights-intra}
\omega_{q_v}^{intra}=\exp(-w\cdot(1 - [{P}_{q_v}^{intra}]_{l'})^{2}),
\end{equation}
where $w$ is the hyper-parameter controlling the degree of penalization for unreliable samples, empirically set to 10 in our experiments. The term $[{P}_{q_v}^{intra}]_{l'}$ is the $l'$-th entry of ${P}_{q_v}^{intra}$, with $l'$ representing the pseudo label of $q_v$ in the visible modality. Consequently, $\omega_{q_v}^{intra}$ increases monotonically with $[{P}_{q_v}^{intra}]_{l'}$. Similarly, for heterogeneous learning, the weight of $q_v$ can be expressed in an analogous manner as:
\begin{equation}\label{weights-inter}
\omega_{q_v}^{inter}=\exp(-w\cdot(1 - [{P}_{q_v}^{inter}]_{T^{V\rightarrow I}[l']})^{2}),
\end{equation}
where $T^{V\rightarrow I}[l']$ denotes the matched cluster in the infrared modality for $q_v$. For samples originating from the infrared modality, the weights in both homogeneous and heterogeneous learning processes can be determined in a similar manner. Ultimately, the overall objective function of our method is expressed as follows:
\begin{equation}\label{Our-learning}
\begin{aligned}
\mathcal{L} &= \mathcal{L}_{homo} + \lambda \mathcal{L}_{heter}\\
&=\frac{1}{|B_V|} \sum_{q_v \in B_V}(\omega_{q_v}^{intra} L^{intra}_{soft}({q}_{v}) + \lambda \omega_{q_v}^{inter} L^{inter}_{soft}({q}_{v}))\\
&+ \frac{1}{|B_I|} \sum_{q_i \in B_I}(\omega_{q_i}^{intra}L^{intra}_{soft}({q}_{i})+ \lambda \omega_{q_i}^{inter} L^{inter}_{soft}({q}_{i})),
\end{aligned}
\end{equation}
where $B_I$ and $B_V$ represent the input batches from the infrared and visible modalities, respectively. The parameter $\lambda$ is a hyper-parameter. In the objective function, the first term corresponds to the homogeneous learning process, while the second term pertains to the heterogeneous learning process. 

\textbf{Discussion.} In this section, we propose a method for generating weights for samples to mitigate the influence of unreliable samples during both the homogeneous and heterogeneous learning stages. Initially, we assign higher weights to easily labeled samples to aid the model in learning general patterns. As training advances, more challenging samples are assigned greater weights if they demonstrate increased consistency with their neighbors. This approach can be considered a form of curriculum learning.

\section{Theoretical Analysis}
\label{analysis}
In this study, we derive soft labels from neighbor information to optimize model parameters both in the homogeneous learning and heterogeneous learning processes. In this part, we aim to investigate the Rademacher generalization bound of our method. For simplicity, we merely consider the binary classification setting, i.e., each sample $x$ is paired with a label $y \in \{0, 1\}$. Our method relies on soft labels derived from a probabilistic distribution, leading to the following loss function formulation: $\ell^{\rm SOFT}(h(x), {y})=(1 - \beta) \cdot \ell(h(x), y) + \beta \cdot \ell(h(x), 1-y)$, where $\ell$ represents a commonly used loss function such as cross-entropy loss and $\beta$ denotes the probability of the irrelevant class. Our method aims to obtain the classifier $h$ by minimizing the following expected risk:
\begin{equation}\label{eq_prove1}
\mathcal{R}(h):=\mathbb{E}_{\left(x, {y}\right) \sim \mathbb{D}}\left[\ell^{\rm SOFT}\left(h\left(x\right), {y}\right)\right],
\end{equation}
where $\mathbb{D}$ is the distribution of the dataset and ${y}$ is the corresponding  generated pseudo label. $\ell$ is the loss function. Due to the limited scale of our dataset, our actual learning objective becomes the empirical risk as follows:
\begin{equation}\label{eq_prove2}
\hat{\mathcal{R}}(h):=\frac{1}{N}\sum\nolimits_{i=1}^{N} \left[\ell^{\rm SOFT}\left(h\left(x_i\right), {y_i}\right)\right],
\end{equation}
where $N$ is the size of the training set. 

{\bf Theorem 1}\cite{wei2022smooth}. With probability at least $1-\delta$, for all $h \in \mathcal{H}$, we have:
\begin{equation}\label{eq_prove3}
\mathcal{R}(h)\leq \hat{\mathcal{R}}(h) + 2\cdot L \cdot \Re(\mathcal{H}) + (1 - 2\beta)(\overline{\ell} - \underline{\ell}) \cdot \sqrt{\frac{log(1/\delta)}{2N}},
\end{equation}
where $\overline{\ell}$ and $\underline{\ell}$ are the upper bound and lower bound of $\ell$ and $\Re$ is the Rademacher complexity. 

{\bf Remark 1.} Theorem 1 offers an upper bound for the gap between expected risk $\mathcal{R}(h)$ and empirical risk $\hat{\mathcal{R}}(h)$. Since $\beta$ is usually set to a small value to ensure the pivot can be classified, it can be regarded as a constant here. When the size of the dataset $N$ is large enough and the Rademacher complexity of the hypothesis space $\Re(\mathcal{H})$ is limited, $\mathcal{R}(h)$ can approach $\hat{\mathcal{R}}(h)$ well with soft labels derived from probabilistic distributions.

\begin{table*}[t]

\renewcommand{\arraystretch}{1.06}
\centering
\setlength{\tabcolsep}{1.6mm}{
\begin{tabular}{ccccccccccccccc}
\hline
\rowcolor{light-gray}   & & & \multicolumn{6}{|c}{SYSU-MM01 dataset} & \multicolumn{6}{|c}{RegDB dataset}  \\
 \cline{4-15}
  \rowcolor{light-gray}  & \multicolumn{2}{c}{} & \multicolumn{3}{|c}{Indoor Search} & \multicolumn{3}{|c}{All search} & \multicolumn{3}{|c}{Infrared to Visible} & \multicolumn{3}{|c}{Visible to Infrared}   \\
  \hline
 \rowcolor{light-gray}  & \multicolumn{1}{|c}{Method} & \multicolumn{1}{|c}{Reference} & \multicolumn{1}{|c}{r1} & \multicolumn{1}{c}{mAP} & \multicolumn{1}{c}{mINP} & \multicolumn{1}{|c}{r1} & \multicolumn{1}{c}{mAP} & \multicolumn{1}{c}{mINP} & \multicolumn{1}{|c}{r1} & \multicolumn{1}{c}{mAP} & \multicolumn{1}{c}{mINP} & \multicolumn{1}{|c}{r1} & \multicolumn{1}{c}{mAP} & \multicolumn{1}{c}{mINP}\\
   \hline
   \hline
    \multirow{7}*{\rotatebox{90}{SVI-ReID}}  & \multicolumn{1}{|c}{AGW } & \multicolumn{1}{|c}{TPAMI-21} & \multicolumn{1}{|c}{54.17} & 62.97 & 59.23 & \multicolumn{1}{|c}{47.50} & 47.65 & 35.30 & \multicolumn{1}{|c}{70.49} & 65.90 & 51.24 & \multicolumn{1}{|c}{70.05} & 66.37 & 50.19  \\
    
    & \multicolumn{1}{|c}{CA} & \multicolumn{1}{|c}{ICCV-21} & \multicolumn{1}{|c}{76.26} & 80.37 & 76.79 & \multicolumn{1}{|c}{69.88} & 66.89 & 53.61 & \multicolumn{1}{|c}{84.75} & 77.82 & 61.56 & \multicolumn{1}{|c}{85.03} & 79.14 & 65.33  \\

  & \multicolumn{1}{|c}{MAUM} & \multicolumn{1}{|c}{CVPR-22} & \multicolumn{1}{|c}{76.97} & 81.94 & - & \multicolumn{1}{|c}{71.68} & 68.79 & - & \multicolumn{1}{|c}{86.95} & 84.34 & - & \multicolumn{1}{|c}{87.87} & 85.09 & -  \\
& \multicolumn{1}{|c}{DFLN-ViT} & \multicolumn{1}{|c}{TMM-22} & \multicolumn{1}{|c}{62.13} & 69.03 & - & \multicolumn{1}{|c}{59.84} & 57.70 & - & \multicolumn{1}{|c}{91.21} & 81.62 & - & \multicolumn{1}{|c}{92.10} & 82.11 & -  \\
 & \multicolumn{1}{|c}{PartMix} & \multicolumn{1}{|c}{CVPR-23} & \multicolumn{1}{|c}{81.52} & 84.38 & - & \multicolumn{1}{|c}{77.78} & 74.62 & -  & \multicolumn{1}{|c}{84.93} & 82.52 & - & \multicolumn{1}{|c}{85.66} & 82.27 & - \\
 & \multicolumn{1}{|c}{MUN} & \multicolumn{1}{|c}{ICCV-23} & \multicolumn{1}{|c}{79.42} & 82.06 & - & \multicolumn{1}{|c}{76.24} & 73.81 & - & \multicolumn{1}{|c}{91.86} & 85.01 & - & \multicolumn{1}{|c}{95.19} & 87.15 & -  \\
& \multicolumn{1}{|c}{SAAI} & \multicolumn{1}{|c}{ICCV-23} & \multicolumn{1}{|c}{83.20} & 88.01 & - & \multicolumn{1}{|c}{75.90} & 77.03 & - & \multicolumn{1}{|c}{92.09} & 92.01 & - & \multicolumn{1}{|c}{91.07} & 91.45 & -  \\
\hline
\hline
\multirow{6}*{\rotatebox{90}{USL-ReID}} & \multicolumn{1}{|c}{SPCL} & \multicolumn{1}{|c}{NeurIPS-20} & \multicolumn{1}{|c}{26.83} & 36.42 & 33.05 & \multicolumn{1}{|c}{18.37} & 19.39 & 10.99 & \multicolumn{1}{|c}{11.70} & 13.56 & 10.09 & \multicolumn{1}{|c}{13.59} & 14.86 & 10.36    \\
& \multicolumn{1}{|c}{MMT} & \multicolumn{1}{|c}{ICLR-20} & \multicolumn{1}{|c}{24.42} & 25.59 & 18.66 & \multicolumn{1}{|c}{25.68} & 26.51 & 19.56 & \multicolumn{1}{|c}{22.79} & 31.50 & 27.66 & \multicolumn{1}{|c}{21.47} & 21.53 & 11.50   \\
& \multicolumn{1}{|c}{ICE} & \multicolumn{1}{|c}{ICCV-21} & \multicolumn{1}{|c}{12.18} & 14.82 & 10.60 & \multicolumn{1}{|c}{12.98} & 15.64 & 11.91 & \multicolumn{1}{|c}{29.81} & 38.35 & 34.32 & \multicolumn{1}{|c}{20.54} & 20.39 & 10.24   \\
& \multicolumn{1}{|c}{CCL} & \multicolumn{1}{|c}{ACCV-22} & \multicolumn{1}{|c}{11.14} & 12.99 & 8.99 & \multicolumn{1}{|c}{11.76} & 13.88 & 9.94 & \multicolumn{1}{|c}{23.33} & 34.01 & 30.88 & \multicolumn{1}{|c}{20.16} & 22.00 & 12.97   \\
& \multicolumn{1}{|c}{PPLR} & \multicolumn{1}{|c}{CVPR-22} & \multicolumn{1}{|c}{8.11} & 9.07 & 5.65 & \multicolumn{1}{|c}{8.93} & 11.14 & 7.89 & \multicolumn{1}{|c}{12.71} & 20.81 & 17.61 & \multicolumn{1}{|c}{11.98} & 12.25 & 4.97   \\
& \multicolumn{1}{|c}{ISE} & \multicolumn{1}{|c}{CVPR-22} & \multicolumn{1}{|c}{10.83} & 13.66 & 10.71 & \multicolumn{1}{|c}{16.12} & 16.99 & 13.24 & \multicolumn{1}{|c}{14.22} & 24.62 & 21.74 & \multicolumn{1}{|c}{20.01} & 18.93 & 8.54   \\
\hline 
\hline
\multirow{10}*{\rotatebox{90}{USL-VI-ReID}} & \multicolumn{1}{|c}{H2H$^{*}$} & \multicolumn{1}{|c}{TIP-21} & \multicolumn{1}{|c}{-} & - & - & \multicolumn{1}{|c}{30.15} & 29.40 & - & \multicolumn{1}{|c}{-} & - & - & \multicolumn{1}{|c}{23.81} & 18.87 & -  \\
& \multicolumn{1}{|c}{OTLA$^*$} & \multicolumn{1}{|c}{ECCV-22} & \multicolumn{1}{|c}{47.4} & 56.8 & - & \multicolumn{1}{|c}{48.2} & 43.9 & - & \multicolumn{1}{|c}{49.6} & 42.8 & - & \multicolumn{1}{|c}{49.9} & 41.8 & -  \\
& \multicolumn{1}{|c}{ADCA} & \multicolumn{1}{|c}{MM-22} & \multicolumn{1}{|c}{50.60} & 59.11 & 55.17 & \multicolumn{1}{|c}{45.51} & 42.73 & 28.29 & \multicolumn{1}{|c}{68.48} & 63.81 & 49.62 & \multicolumn{1}{|c}{67.20} & 64.05 & 52.67  \\
 & \multicolumn{1}{|c}{PGM} & \multicolumn{1}{|c}{CVPR-23} & \multicolumn{1}{|c}{56.23} & 62.74 & {58.13} & \multicolumn{1}{|c}{{57.27}} & {51.78} & {34.96} & \multicolumn{1}{|c}{69.85} & 65.17 & - & \multicolumn{1}{|c}{69.48} & 65.41 & -  \\

   & \multicolumn{1}{|c}{DOTLA$^{*}$} & \multicolumn{1}{|c}{MM-23} & \multicolumn{1}{|c}{53.47} & 61.73 & 57.35 & \multicolumn{1}{|c}{50.36} & 47.36 & 32.40 & \multicolumn{1}{|c}{\underline{82.91}} & 74.97 & 58.60 & \multicolumn{1}{|c}{\underline{85.63}} & 76.71 & 61.58  \\
  & \multicolumn{1}{|c}{MBCCM } & \multicolumn{1}{|c}{MM-23} & \multicolumn{1}{|c}{55.21} & 61.98 & 57.13 & \multicolumn{1}{|c}{53.14} & 48.16 & 32.41 & \multicolumn{1}{|c}{82.82} & \underline{76.74} & \underline{61.73} & \multicolumn{1}{|c}{83.79} & {77.87} & \underline{65.04}  \\
 & \multicolumn{1}{|c}{CCLNet} & \multicolumn{1}{|c}{MM-23} & \multicolumn{1}{|c}{{56.68}} & {65.12} & - & \multicolumn{1}{|c}{54.03} & 50.19 & - & \multicolumn{1}{|c}{70.17} & 66.66 & - & \multicolumn{1}{|c}{69.94} & 65.53 & -  \\
  & \multicolumn{1}{|c}{GUR$^\dag$} & \multicolumn{1}{|c}{ICCV-23} & \multicolumn{1}{|c}{\underline{64.22}} & \underline{69.49} & \underline{64.81} & \multicolumn{1}{|c}{\underline{60.95}} & \underline{56.99} & \underline{41.85} & \multicolumn{1}{|c}{75.00} & 69.94 & 56.21 & \multicolumn{1}{|c}{73.91} & 70.23 & 58.88  \\
  & \multicolumn{1}{|c}{SCA-RCP} & \multicolumn{1}{|c}{TKDE-24} & \multicolumn{1}{|c}{56.77} & 64.19 & 59.25 & \multicolumn{1}{|c}{51.41} & 48.52 & 33.56 & \multicolumn{1}{|c}{82.41} & 75.73 & - & \multicolumn{1}{|c}{85.59} & \underline{79.12} & -  \\
& \multicolumn{1}{|c}{Ours} & \multicolumn{1}{|c}{-} & \multicolumn{1}{|c}{\textbf{67.04}} & \textbf{73.08} & \textbf{69.42} & \multicolumn{1}{|c}{\textbf{61.81}} & \textbf{58.92} & \textbf{45.01} & \multicolumn{1}{|c}{\textbf{88.17}} & \textbf{81.11} & \textbf{66.05} & \multicolumn{1}{|c}{\textbf{88.75}} & \textbf{82.14} & \textbf{68.75}  \\
\hline   
\end{tabular}
}
\caption{Experimental results (\%) of our method and SOTA methods on the SYSU-MM01 and RegDB datasets under different settings. $*$ means the model is pre-trained on an extra labeled visible dataset. GUR$^\dag$ deontes results without camera information. }
\label{tab:1}
\end{table*}

\begin{table*}[htbp]

\renewcommand{\arraystretch}{1.0}
\centering
\setlength{\tabcolsep}{1.43mm}{
\begin{tabular}{ccccccccccccccccc}
\hline
\multirow{3}*{Index} & \multicolumn{3}{|c}{{\multirow{2}{*}{Components}}} & \multicolumn{6}{|c}{SYSU-MM01 Settings} & \multicolumn{6}{|c}{RegDB Settings}  \\
\cline{5-16}
& \multicolumn{3}{|c}{} & \multicolumn{3}{|c}{Indoor Search} & \multicolumn{3}{|c}{All search} & \multicolumn{3}{|c}{Infrared-to-Visible} & \multicolumn{3}{|c}{Visible-to-Infrared} \\
\cline{2-16}
& \multicolumn{1}{|c}{Baseline} & \multicolumn{1}{c}{N-ULC} & \multicolumn{1}{c}{N-DW} & \multicolumn{1}{|c}{r1} & mAP & mINP & \multicolumn{1}{|c}{r1} & mAP & mINP & \multicolumn{1}{|c}{r1} & mAP & mINP & \multicolumn{1}{|c}{r1} & mAP & mINP\\
\hline
1 & \multicolumn{1}{|c}{\Checkmark} & \multicolumn{1}{c}{} & \multicolumn{1}{c}{} & \multicolumn{1}{|c}{58.80} & 65.35 & 60.58 & \multicolumn{1}{|c}{54.14} & 50.37 & 33.56 & \multicolumn{1}{|c}{77.44} & 66.47 & 45.40 & \multicolumn{1}{|c}{77.92} & 67.51 & 48.43\\
2 & \multicolumn{1}{|c}{\Checkmark} & \multicolumn{1}{c}{\Checkmark} & \multicolumn{1}{c}{} & \multicolumn{1}{|c}{61.14} & 67.15 & 62.69 & \multicolumn{1}{|c}{55.87} & 51.79 & 35.71 & \multicolumn{1}{|c}{79.29} & 70.45 & 52.19 & \multicolumn{1}{|c}{80.60} & 71.72 & 54.64\\
3 & \multicolumn{1}{|c}{\Checkmark} & \multicolumn{1}{c}{} & \multicolumn{1}{c}{\Checkmark} & \multicolumn{1}{|c}{62.04} & 69.06 & 65.16 & \multicolumn{1}{|c}{58.54} & 55.92 & 41.79 & \multicolumn{1}{|c}{87.82} & \textbf{81.80} & \textbf{67.66} & \multicolumn{1}{|c}{87.88} & \textbf{82.24} & \textbf{70.13}\\
4 & \multicolumn{1}{|c}{\Checkmark} & \multicolumn{1}{c}{\Checkmark} & \multicolumn{1}{c}{\Checkmark} & \multicolumn{1}{|c}{\textbf{67.04}} & \textbf{73.08} & \textbf{69.42} & \multicolumn{1}{|c}{\textbf{61.81}} & \textbf{58.92} & \textbf{45.01} & \multicolumn{1}{|c}{\textbf{88.17}} & {81.11} &{66.05} & \multicolumn{1}{|c}{\textbf{88.75}} & {82.14} & {68.75} \\
\hline

\end{tabular}
}
\caption{Ablation study on the SYSU-MM01 and RegDB datasets (\%). }
\label{ablation}
\end{table*}

\section{Experiment}
\subsection{Datasets and Evaluation Protocols}
Our experiments are conducted on two public datasets: RegDB \cite{nguyen2017person} and SYSU-MM01 \cite{wu2017rgb}. To ensure fair comparison, we use mean average precision (mAP), Cumulative Matching Characteristics (CMC), and mean Inverse Negative Penalty (mINP) \cite{ye2021deep} as evaluation metrics, which are commonly employed in existing research \cite{yang2022augmented,wu2023unsupervised}.

SYSU-MM01 is a large-scale VI-ReID dataset collected from 4 visible and 2 infrared cameras. The training set includes 395 identities with 22,258 visible images and 11,909 infrared images, while the testing set includes 96 identities. 

RegDB is another dataset acquired from one visible and one infrared camera. It contains 412 identities, each identity includes 10 visible and 10 infrared images. In line with previous studies \cite{yang2022augmented,wu2023unsupervised}, we perform 10 experiments on this dataset and report the average performance as the final results.

\subsection{Implementation Details}
Our method is implemented in the PyTorch platform. Following existing works \cite{yang2022augmented,cheng2023efficient,yang2023towards,li2024instance}, we utilize the ImageNet-pretrained ResNet50 \cite{he2016deep} as the backbone. In each mini-batch, we select 16 identities per modality, with each identity comprising 16 instances. Input images are resized to 288 $\times$ 144 pixels. Standard data augmentation techniques, including random flipping, random cropping, and random erasing, are applied. Pseudo labels are generated using the DBSCAN clustering algorithm, with a maximum distance of 0.2 for the RegDB dataset and 0.6 for the SYSU-MM01 dataset. Hyper-parameters $\mu$ and $\lambda$ are set to 0.7 and 3, respectively, while the number of neighbors $k$ is set to 20 and 30 for the RegDB and SYSU-MM01 datasets, respectively. All other experimental settings follow those of previous works \cite{yang2022augmented,wu2023unsupervised}.

\subsection{Comparison with SOTA Methods}
To assess the effectiveness of our method, we compare it against state-of-the-art approaches across three ReID settings: supervised visible-infrared person ReID (SVI-ReID), unsupervised single-modality person ReID (USL-ReID), and unsupervised visible-infrared person ReID (USL-VI-ReID). The results are presented in Table~\ref{tab:1}.

\textbf{Comparison with SVI-ReID methods.} 
We compare our method with several recent SVI-ReID approaches, including AGW \cite{ye2021deep}, CA \cite{ye2021channel}, MAUM \cite{liu2022learning}, DFLN-ViT \cite{zhao2022spatial}, PartMix \cite{Kim_2023_CVPR}, MUN \cite{yu2023modality}, 
and SAAI \cite{fang2023visible}. Despite these methods benefiting from precise manual annotations, our method achieves comparable performance to some of them, such as AGW and DFLN-ViT.

\textbf{Comparison with USL-ReID methods.} 
We compare our method with recent state-of-the-art USL-ReID techniques, including SPCL \cite{ge2020self}, MMT \cite{ge2020mutual}, CCL \cite{dai2021cluster}, ICE \cite{chen2021ice}, PPLR \cite{cho2022part}, and ISE \cite{zhang2022implicit}. These approaches are generally constrained by their focus on USL-ReID, which limits their ability to address severe modality discrepancies.

\textbf{Comparison with USL-VI-ReID methods.} 
We also evaluate our method against advanced USL-VI-ReID approaches, such as H2H \cite{liang2021homogeneous}, OTLA \cite{wang2022optimal}, ADCA \cite{yang2022augmented}, TAA \cite{yang2023translation}, PGM \cite{wu2023unsupervised}, CCLNet \cite{chen2023unveiling}, MBCCM \cite{cheng2023efficient}, DOTLA \cite{cheng2023unsupervised}, GUR \cite{yang2023towards}, and SCA-RCP \cite{li2024inter}. Our method outperforms all of these approaches on the tested datasets, likely because it effectively addresses universal label noise in homogeneous and heterogeneous learning contexts at both the pseudo label and sample levels, thus demonstrating superior efficiency.

\subsection{Ablation Study}
In Table \ref{ablation}, we present experiments conducted on the SYSU-MM01 and RegDB datasets to evaluate the components of our method, specifically the Neighbor-Guided Label Universal Calibration (N-ULC) and Neighbor-Guided Dynamic Weighting (N-DW) modules. The results indicate that applying the N-ULC and N-DW modules individually yields substantial improvements on both datasets, highlighting the effectiveness of these components. When both modules are combined, our method achieves optimal performance on the SYSU-MM01 dataset. However, improvements on the RegDB dataset are more limited. This limitation is likely due to the already high performance of the N-DW module on RegDB and the partial overlap in functionality between the modules, which restricts further gains in performance. 

\begin{figure}[htbp]
\centering
\subfigure[Impact of $k$ on SYSU-MM01 under all search setting.]{\includegraphics[width=0.233\textwidth, height=3.0cm]{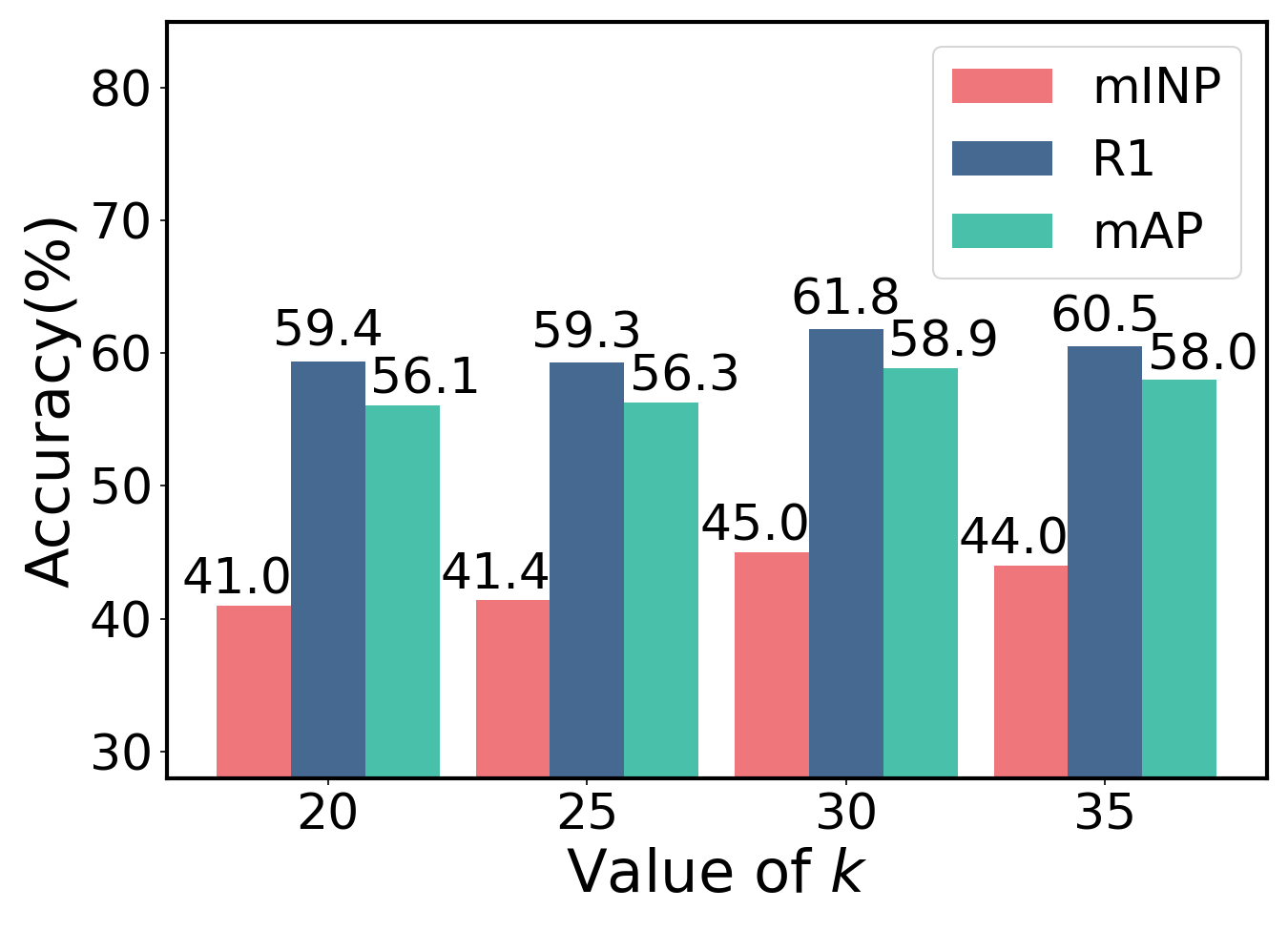}}\hspace{0mm}
\subfigure[Impact of $k$ on RegDB under visible-to-infrared setting.]{\includegraphics[width=0.233\textwidth, height=3.0cm]{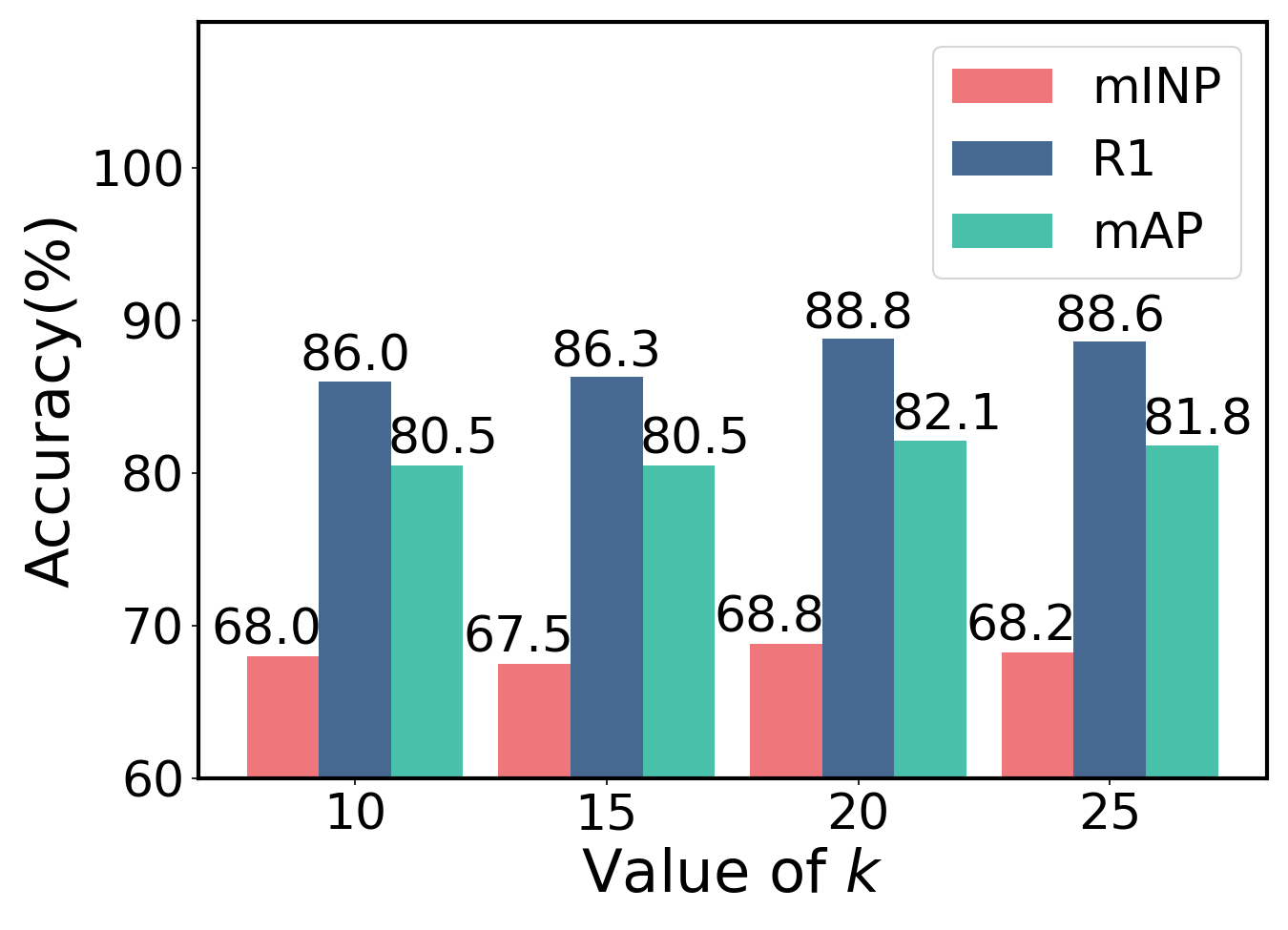}}
\caption{Impact of hyper-parameter $k$ on different datasets.}
\label{neighbor}
\end{figure}

\subsection{Hyper-parameter Analysis}
Our method relies on information from nearest neighbors, making the nearest neighbor number $k$ in Eq. \eqref{correlation} a crucial factor in its performance. Figure \ref{neighbor} presents experiments conducted on various datasets to assess the impact of this hyper-parameter. The results indicate that the performance of our method remains relatively stable across different values of $k$. For optimal results, $k$ is set to 30 for the SYSU-MM01 dataset and 20 for the RegDB dataset.

\subsection{Visualization}
\begin{figure}[t]
\centering
\subfigure[Ours on RegDB]{\includegraphics[width=3.9cm, height=2.8cm]{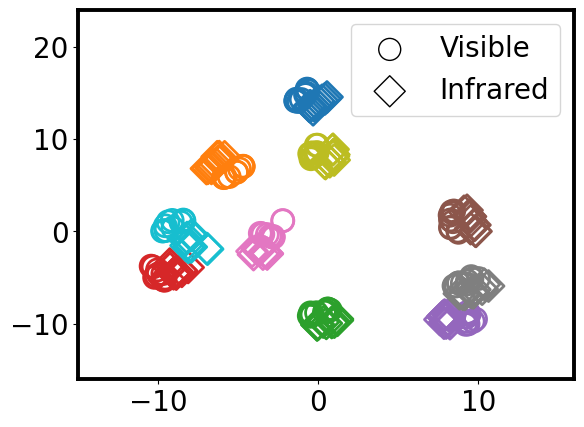}}
\subfigure[{PGM on RegDB}]{\includegraphics[width=3.9cm, height=2.8cm]{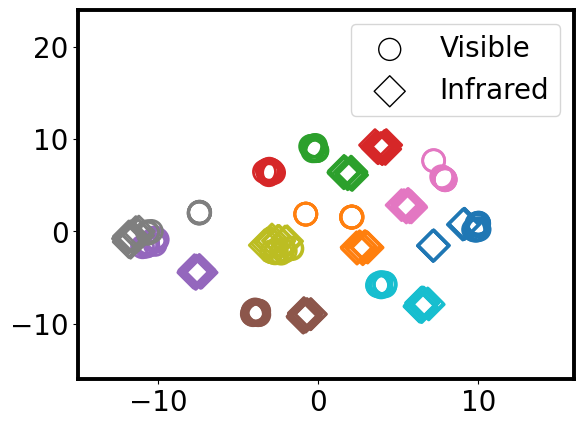}}\hspace{0mm}
\subfigure[Ours on SYSU-MM01]{\includegraphics[width=3.9cm, height=2.8cm]{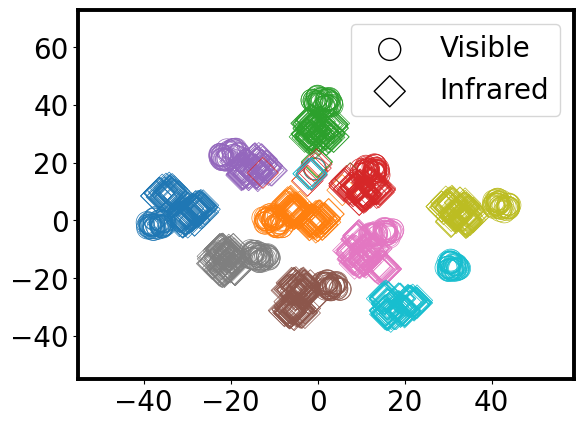}}
\subfigure[PGM on SYSU-MM01]{\includegraphics[width=3.9cm, height=2.8cm]{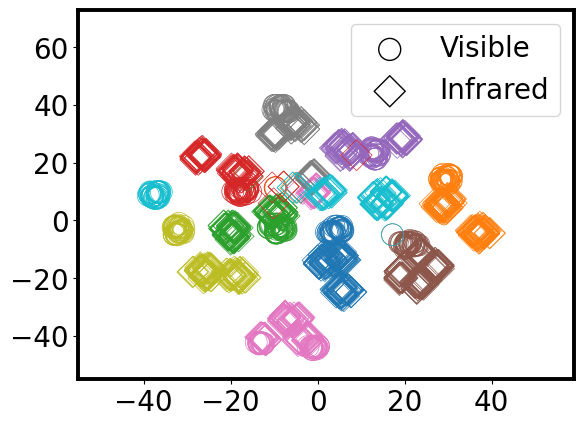}}\hspace{0mm}
\caption{T-SNE visualization of features learned by PGM and our method on a subset of RegDB and SYSU-MM01 datasets. Different colors
represent different identities.}
\label{tsne}
\end{figure}

To further evaluate the effectiveness of our method in learning modality-shareable feature representations, we employ t-SNE \cite{van2008visualizing} to visualize the features learned by both our method and PGM on the SYSU-MM01 and RegDB datasets, as shown in Fig. \ref{tsne}. The results reveal that PGM often separates images of the same person across different modalities into distinct clusters, whereas our method generates more compact feature representations for the same individual. In contrast, PGM tends to confuse more identities and blend features from different individuals compared with our method.

\section{Conclusion}
In this paper, we present a simple yet effective framework for USL-VI-ReID that addresses universal label noise through the use of neighbor information. To mitigate label noise, we introduce the neighbor-guided universal label calibration module, which refines explicit hard pseudo labels by leveraging information from neighbors in both homogeneous and heterogeneous spaces. Additionally, to enhance training stability, we propose the neighbor-guided dynamic weighting module, which reduces the impact of unreliable samples within and across modalities. Extensive experiments conducted on the SYSU-MM01 and RegDB datasets demonstrate the effectiveness of our proposed method, despite its inherent simplicity.

\section{Acknowledgments}
This work was supported by the National Natural Science Foundation of China (No. 62376282)

\bibliography{aaai25}

\newpage

In the supplementary material, we present additional experimental results and provide further details on our experimental procedures. 

 \section{More Implementation Details}
Our framework is implemented using PyTorch, with training conducted on four NVIDIA A100 GPUs and testing performed on a single GPU. For input images, we apply various data augmentation techniques, including random horizontal flipping, padding, random grayscale conversion, channel augmentation \cite{ye2021channel}, random cropping, and random erasing. Following existing works \cite{yang2022augmented,wu2023unsupervised}, our method adopts the two-stage training approach. To train the model, Adam optimizer with weight decay 5e-4 is adopted. For each stage, we set the initial learning rate as 3.5e-4, and reduce it every 20 epochs for a total 50 epochs.

During training, we use the DBSCAN clustering algorithm \cite{ester1996density} to generate pseudo labels within each modality. For DBSCAN, we set the minimum number of neighbors to 4 across all datasets. The maximum distance parameter is set to 0.6 for the SYSU-MM01 dataset and 0.2 for the RegDB dataset, aligning with the values used in similar studies. Following \cite{wu2023unsupervised}, the momentum factor $m$ in the PGM module is set to 0.2 and the temperature hyper-parameter $\tau$ is set to 0.05. Hyper-parameters $\mu$ and $\lambda$ are set to 0.7 and 3, respectively, while the number of neighbors $k$ is set to 20 and 30 for the RegDB and SYSU-MM01 datasets, respectively.

\begin{figure}[htbp]
\centering
\subfigure[Impact of $\lambda$ on SYSU-MM01 under all search setting.]{\includegraphics[width=0.233\textwidth, height=3.1cm]{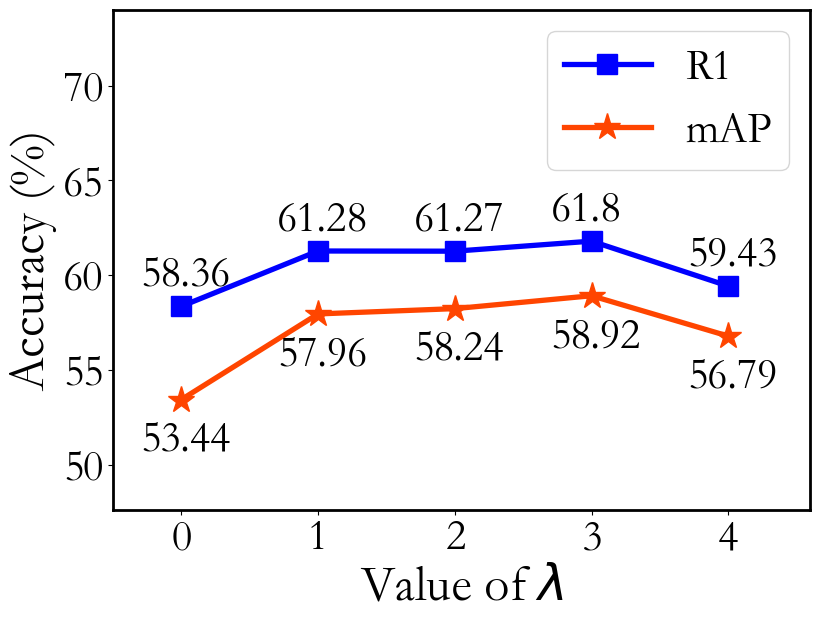}}\hspace{0mm}
\subfigure[Impact of $\mu$ on SYSU-MM01 under all search setting.]{\includegraphics[width=0.233\textwidth, height=3.1cm]{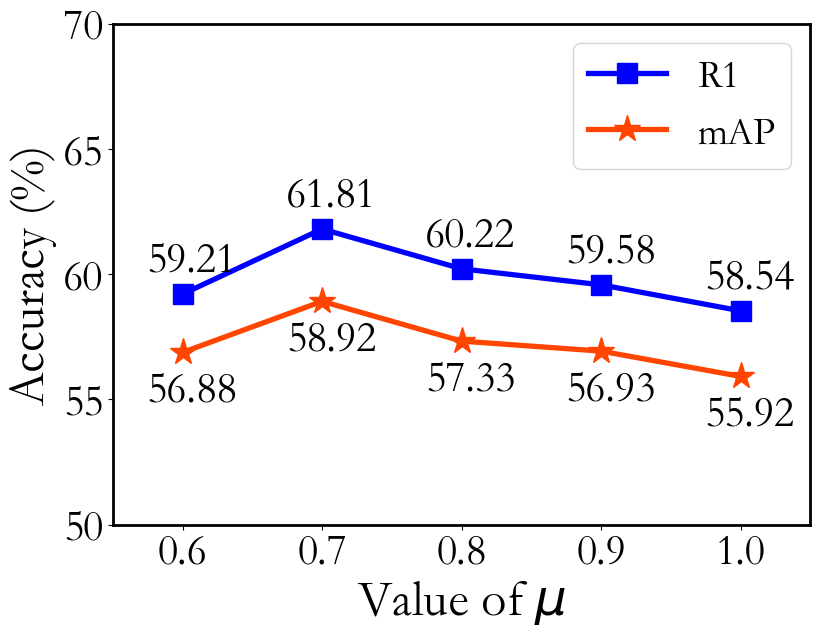}}
\caption{Impact of hyper-parameters $\lambda$ and $\mu$ on SYSU-MM01 dataset.}
\label{mu_lambda}
\end{figure}

\section{More Experimental Results}
\subsection{Influence of the hyper-parameter $\lambda$}
Our method involves both homogeneous learning and heterogeneous learning processes. In Eq. (13), the hyper-parameter $\lambda$ is a crucial factor which balances the weights of homogeneous learning and heterogeneous learning. When $\lambda$ increases, the model will focus more on the association between different modalities while neglecting the intrinsic patterns within the modality, and vice versa. To obtain the optimal value of $\lambda$, we conduct experiments on SYSU-MM01 dataset with different values of this hyper-parameter and results are shown in Fig. \ref{mu_lambda}(a). From the results, we can find that the performance of the model is relatively robust against different values of $\lambda$. For simplicity, we set $\lambda$ to 3.0 in both SYSU-MM01 and RegDB datasets.

\subsection{Influence of the hyper-parameter $\mu$}
The hyperparameter $\mu$ in the N-ULC module controls the calibration strength of pseudo labels and is crucial for the module’s performance. If $\mu$ is set too small, the calibration of the original pseudo labels will be insufficient. On the other hand, if $\mu$ is set too large, the calibration may override the original pseudo label information, hindering the model’s ability to learn discriminative feature representations. Fig. \ref{mu_lambda}(b) illustrates the model’s performance on the SYSU-MM01 dataset for different values of $\mu$. The results support these observations: when $\mu$ is set to 0.6 or 1.0, the model's performance drops significantly. However, with $\mu$ set to 0.7, the model achieves the best performance. For consistency, we set $\mu$ to 0.7 across all datasets.

\subsection{Influence of the hyper-parameter $w$}
The hyperparameter $w$ in the N-DW module controls the penalty applied to unreliable samples. If $w$ is set too small, the weight differences between samples become negligible, making it hard to distinguish between them effectively. On the other hand, if $w$ is set too large, the module suppresses too many samples, making it insufficient for model training. Table \ref{w} shows the model's performance on the SYSU-MM01 dataset with different values of $w$. The model achieves the best performance when $w$ is set to 10. For consistency, $w$ is set to 10 across all datasets in the paper. 

\begin{table}[htbp]

\renewcommand{\arraystretch}{1.1}
\centering
\setlength{\tabcolsep}{2.1mm}{
\begin{tabular}{c|ccccccc}
\hline
\multirow{2}*{$w$} & \multicolumn{3}{|c}{Indoor Search} & \multicolumn{3}{|c}{All Search} \\
\cline{2-7}
 & \multicolumn{1}{|c}{r1} & mAP & mINP & \multicolumn{1}{|c}{r1} & mAP & mINP \\
\hline
0.0 & \multicolumn{1}{|c}{61.15} & 67.15 & 62.69 & \multicolumn{1}{|c}{55.87} & 51.79 & 35.71 \\
2.5 & \multicolumn{1}{|c}{61.35} & 68.10 & 64.13 & \multicolumn{1}{|c}{58.64} & 55.53 & 40.84 \\
5.0 & \multicolumn{1}{|c}{65.70} & 71.68 & 67.74 & \multicolumn{1}{|c}{60.94} & 58.00 & 43.96 \\
7.5 & \multicolumn{1}{|c}{63.07} & 69.85 & 66.01 & \multicolumn{1}{|c}{59.56} & 55.89 & 41.19 \\
10.0 & \multicolumn{1}{|c}{\textbf{67.04}} & \textbf{73.08} & \textbf{69.42} & \multicolumn{1}{|c}{\textbf{61.81}} & \textbf{58.92} & \textbf{45.01} \\
12.5 & \multicolumn{1}{|c}{63.79} & 70.79 & 67.19 & \multicolumn{1}{|c}{59.36} & 56.64 & 42.43 \\
\hline

\end{tabular}
}
\caption{Impact of hyper-parameter $w$ on SYSU-MM01 dataset.}
\label{w}
\end{table}

\end{document}